\pdfoutput=1

\documentclass[11pt]{article}

\usepackage[]{EMNLP2022}

\usepackage{times}
\usepackage{latexsym}

\usepackage[T1]{fontenc}

\usepackage[utf8]{inputenc}

\usepackage{microtype}

\usepackage{inconsolata}

\usepackage{subfigure}
\usepackage{times}
\usepackage{threeparttable}
\usepackage{setspace}
\usepackage{multirow}
\usepackage{graphicx}
\usepackage{amsmath}
\usepackage{booktabs}
\usepackage{tabularx}
\usepackage{paralist}
\usepackage{bbding}
\usepackage{amssymb}

\title{\textsc{WR-One2Set}: Towards Well-Calibrated Keyphrase Generation}

\author{Binbin Xie$^{1,3}$, Xiangpeng Wei$^{2}$, Baosong Yang$^{2}$, Huan Lin$^{2}$, Jun Xie$^{2}$, \\
{\bf Xiaoli Wang$^{3}$, Min Zhang$^{4}$ \and Jinsong Su$^{1,3}$\thanks{* Corresponding author.}}\\
        $^1${School of Informatics, Xiamen University, China\ \ }  $^2$Alibaba Group, China \\
        $^3$Key Laboratory of Digital Protection and Intelligent Processing of Intangible\\ Cultural Heritage of Fujian and Taiwan, Ministry of Culture and Tourism, China \\
        $^4$Soochow University, China \\
        \texttt{xdblb@stu.xmu.edu.cn\ } \texttt{pemywei@gmail.com}\\
        \texttt{yangbaosong.ybs@alibaba-inc.com\ } \texttt{sjs@xmu.edu.cn} \\}

\begin{document}
\maketitle
\begin{abstract}
Keyphrase generation aims to automatically generate short phrases summarizing an input document.
The recently emerged \textsc{One2Set} paradigm \cite{ye-etal-2021-one2set} generates keyphrases as a set and has achieved competitive performance.  Nevertheless, we observe serious calibration errors outputted by \textsc{One2Set},
especially in the \textit{over-estimation} of $\varnothing$ token (means ``no corresponding keyphrase'').
In this paper,
we deeply analyze this limitation and identify two main reasons behind:
1) the parallel generation has to introduce excessive $\varnothing$ as padding tokens into training instances;
and
2) the training mechanism assigning target to each slot is unstable and further aggravates the $\varnothing$ token over-estimation. To make the model well-calibrated, we propose \textsc{WR-One2Set} which extends
\textsc{One2Set} with an adaptive instance-level cost \textbf{W}eighting strategy and a target \textbf{R}e-assignment mechanism.
The former dynamically penalizes the over-estimated slots for different instances thus smoothing the uneven training distribution.
The latter refines the original inappropriate assignment and reduces the supervisory signals of over-estimated slots.
Experimental results on commonly-used datasets demonstrate the effectiveness and generality of our proposed paradigm.
\end{abstract}

\section{Introduction}
Keyphrases
are short phrases fully encoding the main information of a given document.
They can not only facilitate readers to quickly understand the document, but also provide useful information to many downstream  tasks,
including document classification \cite{DBLP:conf/acl/HulthM06}, summarization \cite{DBLP:conf/acl/WangC13},
etc.

\begin{figure}[t]
\centering
\small
\includegraphics[width=0.475\textwidth,
trim=10 100 700 190,clip]{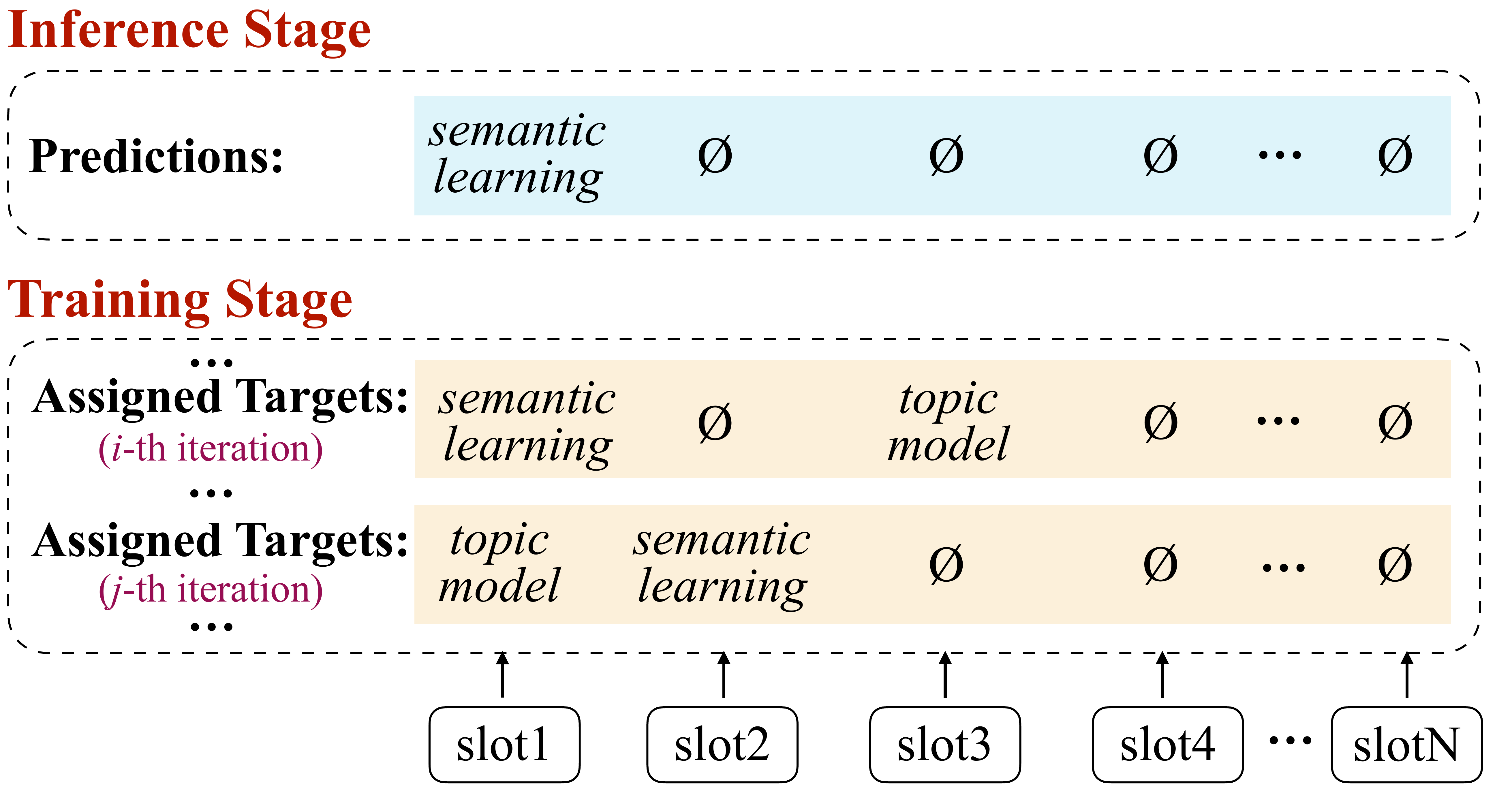}
\setlength{\abovecaptionskip}{5pt}
\caption{
An example of \textsc{One2Set} paradigm at training and inference stages.
``\textrm{Assigned Targets ($*$-th iteration)}'' represents the multiple feasible target permutations generated by $K$-step target assignment mechanism at different training iterations.
In this case,
both ``slot2'' and ``slot3'' are expected to generate keyphrases.
However, they often use $\varnothing$ token as  supervisory signals,
and thus over-estimate and output $\varnothing$ token.
}
\label{fig:paradigm_one2set}
\end{figure}

With the rapid development of deep learning,
\emph{keyphrase generation}~\cite{meng-etal-2017-deep} has attracted increasing attention due to its ability to produce phrases that even do not match any contiguous subsequence of
the source document.\footnote{An example is shown in Appendix, Table \ref{table:present-cn-main-performance}.}
Dominant models of keyphrase generation are constructed under three paradigms:
\textsc{One2One} \cite{meng-etal-2017-deep},
\textsc{One2Seq} \cite{yuan-etal-2020-one} and
\textsc{One2Set} \cite{ye-etal-2021-one2set}.
Among these paradigms,
\textsc{One2Set} exhibits the state-of-the-art (SOTA) performance.
As illustrated in Figure~\ref{fig:paradigm_one2set},
it considers keyphrase generation as a set generation task.
After padding keyphrases to a fixed number with special token $\varnothing$,
they define multiple slots
that individually generate each keyphrase in parallel.
During training, each slot is assigned with a keyphrase or $\varnothing$ token\footnote{In this work, we define that the keyphrase can not be a $\varnothing$ token.}
via a $K$-step target assignment mechanism.
Specifically,
the model first generates $K$ tokens from each slot and then
determines the optimal target assignment using a bipartite matching algorithm \cite{DBLP:books/daglib/p/Kuhn10}.
The superiority of \textsc{One2Set} stems from its conditional independence, that is, the prediction distribution of each slot depends only on the given document other than the order of keyphrases like \textsc{One2Seq}. This is more compatible with the unordered property of keyphrases and decreases the difficulty of the model training \cite{ye-etal-2021-one2set}.

Despite of its success, we observe serious over-estimation problem
on $\varnothing$ token, which significantly affects the generation quality. 
For example,
in Figure \ref{fig:paradigm_one2set},
both ``slot2'' and ``slot3'' are expected to generate keyphrases,
but $\varnothing$ token is over-confidently given.
Two questions naturally arise:
1) \textit{what are reasons behind the over-estimation problem in \textsc{One2Set}?}
and 2) \textit{how can we alleviate them?}

In order to answer the first question,
we conduct extensive analyses, and conclude two reasons.
Firstly,
the over-estimation is a by-product inherently carried by the parallel generation.
More concretely,
excessive $\varnothing$ tokens have been introduced as the padding tokens and served as supervisory signals in training data.
The unbalanced data and the lack of dependency among slots leads each slot to learn to commonly generate $\varnothing$ token.
Secondly,
the $K$-step target assignment mechanism provides multiple feasible target permutations that are assigned to slots.
As shown in Figure \ref{fig:paradigm_one2set},
the targets of the given document can be assigned in different permutation at each training iteration,
which further increases the probability of $\varnothing$ token to be assigned as supervisory signal for each slot, thus exacerbating the over-estimation problem.
Both problems make the learned probabilities of the assigned targets deviate from its ground truth likelihood,
finally constructing a miscalibrated model.

Consequently,
we approach the above problems from the calibration perspective and propose two strategies that extend  \textsc{One2Set} to \textsc{WR-One2Set}.
Specifically,
an \textit{adaptive instance-level cost weighting} is first introduced to
penalize the over-estimated slots of different instances.
According to the seriousness of the issue,
instances are rendered different weights,
therefore dynamically balancing the model training.
Besides,
we propose a \textit{target re-assignment} mechanism to refine the original inappropriate assignment and reduce the supervisory signals of $\varnothing$ token.
In particular,
we re-assign targets for the slots potentially generating fresh keyphrases but being pre-assigned with $\varnothing$ token.
In these ways,
\textsc{WR-One2Set} is encouraged to produce well-calibrated probabilities on keyphrase generation.
Overall, major contributions of our work are three-fold:
\begin{itemize}
\setlength{\itemsep}{0pt}
\setlength{\parsep}{0pt}
\setlength{\parskip}{0pt}
\item Through in-depth analyses,
we point out that the advanced keyphrase generation architecture \textsc{One2Set} suffers from the $\varnothing$ token over-estimation,
which is inherently caused by its parallism and the target assignment mechanism.  
\item We propose \textsc{WR-One2Set} which enhances the original framework with two effective strategies to calibrate the over-estimation problem from the training perspective.
\item Extensive experiments on five widely-used datasets reveal the universal-effectiveness of our model. 
\item {We release our code at
\url{https://github.com/DeepLearnXMU/WR-One2Set}.}
\end{itemize}

\section{Related Work}
Early studies mainly focus on automatic keyphrase extraction \cite{hulth-2003-improved,mihalcea-tarau-2004-textrank,DBLP:conf/icadl/NguyenK07,DBLP:conf/aaai/WanX08},
which aims to directly extract keyphrases from the input document.
Recently,
with the rapid development of deep learning,
neural network-based models have been widely used in keyphrase generation.
Typically,
these models are based on an attentional encoder-decoder framework equipped with copy mechanism,
which is able to generate both present and absent keyphrases \cite{meng-etal-2017-deep}.
Generally,
these models are constructed under the following paradigms:
1) \textsc{One2One} \cite{meng-etal-2017-deep,chen-etal-2019-integrated,DBLP:conf/aaai/ChenGZKL19}.
Under this paradigm,
the input document is paired with each target keyphrase to form an independent training instance for model training.
During inference,
the models are encouraged to produce multiple keyphrases via beam search.
2) \textsc{One2Seq} \cite{DBLP:conf/acl/ChanCWK19,yuan-etal-2020-one,chen-etal-2020-exclusive,DBLP:conf/acl/WuLLNCZW21}.
It considers keyphrase generation as a sequence generation task,
where different keyphrases are concatenated into a sequence in a predefined order.
In this way,
the semantic dependence between keyphrases can be exploited to benefit keyphrase generation.
3) \textsc{One2Set} \cite{ye-etal-2021-one2set}.
Unlike \textsc{One2Seq},
this paradigm considers the keyphrases as a set,
which can be predicted from slots 
in a parallel manner and partial target matching algorithm.

Considering that \textsc{One2One} neglects the correlation among keyphrases,
the most popular paradigm \textsc{One2Seq} exploits the correlation by pre-defining the keyphrases order for model training and inference.
Nevertheless,
\textsc{One2Seq} is the opposite of the flexible and unordered properties of keyphrases,
increasing the difficulty of the model training.
Due to the parallelism and the conditional independence,
\textsc{One2Set} attracts much attention in the keyphrase generation community,
and achieves the SOTA performance.
As this method has just been put forward,
it is inevitable to exist imperfections.
Hence, we are committed to analyses and further optimizing this framework.
To the best of our knowledge,
this is the first attempt to improve \textsc{One2Set}.

\section{Background}\label{section:background}
Here,
we briefly introduce
\textsc{SetTrans} \cite{ye-etal-2021-one2set},
which is based on the \textsc{One2Set} paradigm.
It is a Transformer-based, semi-autoregressive model.
Typically, it introduces $N$ slots, each of which introduces a learnable control code as the additional decoder input,
to generate keyphrases or $\varnothing$ tokens in parallel.
Its training involves two stages:
1) a $K$-step target assignment mechanism is firstly used to determine the correspondence between each prediction and target,
and then
2) a  new training objective is introduced to optimize the whole model.
It contains two set losses to separately deal with two kinds of keyphrases:
\emph{present keyphrases} appearing in the input document,
and \emph{absent keyphrases} that do not match any contiguous subsequence of the document.

\paragraph{$K$-Step Target Assignment}
At this stage,
the model predicts $K$ tokens from each slot,
where the predicted probability distributions are also collected.
Then,
an optimal assignment $m$ between predictions and targets can be found by a bipartite matching algorithm \cite{DBLP:books/daglib/p/Kuhn10}:
\begin{equation}
\setlength{\abovedisplayskip}{3pt}
{m = \mathop{\arg\min}_{m \in M(N)} \sum_{i=1}^N \mathcal{C}_{\text{match}}({y^{m(i)}}, {\textbf{P}^i})\label{formulation:1}},
\setlength{\belowdisplayskip}{3pt}
\end{equation}
where $M(N)$ denotes a set of all $N$-length target index permutations,
and the optimal permutation $m$ can be considered as a mapping function from the slot $i$ to the target index $m(i)$.\footnote{Please note that instead of following \citet{ye-etal-2021-one2set} to use the function $\pi(i')$ mapping the target index $i'$ to the slot index $\pi(i')$,
we use $m(i)$ that is the inverse function of $\pi(i)$,
so as to facilitate subsequent descriptions.}
$\mathcal{C}_{\text{match}}(y^{m(i)}, \textbf{P}^{i})$ is a pair-wise matching loss between the target $y^{m(i)}$ and the predicted probability distributions ${\textbf{P}}^{i}$ of the slot $i$.
Note that the set of targets are also padded to size $N$ with $\varnothing$ tokens.

\paragraph{Model Optimization with Set Losses}
During the second stage,
the model is trained with the sum of two set losses.
Concretely,
slots are equally split into two sets,
dealing with the generations of present and absent keyphrases, respectively.
Next,
the above target assignment is performed on these two sets separately,
forming a mapping $m^p$ for present keyphrases,
and a mapping $m^a$ for absent keyphrases.
Finally,
the training objective becomes\begin{equation}
\setlength{\abovedisplayskip}{3pt}
\footnotesize
{\mathcal{L}(\theta) = -{\Bigg[\sum_{i=1}^{\frac{N}{2}}}{\mathcal{L}^p(\theta, y^{m^{p}(i)})+ \sum_{i=\frac{N}{2}+1}^{N}}{\mathcal{L}^a(\theta, y^{m^{a}(i)})\Bigg]}
}\label{formulation:2}
\setlength{\belowdisplayskip}{0pt}
\end{equation}
\begin{equation}
\footnotesize
\setlength{\abovedisplayskip}{0pt}
{\mathcal{L}^p(\theta, z)=\begin{cases}
\lambda_{pre}\cdot\sum_{t=1}^{|z|}\log\hat{p}_t^i(z_t)&\mbox{if $z$=$\varnothing$}\\
\sum_{t=1}^{|z|}\log\hat{p}_t^i(z_t)&\mbox{otherwise}
\end{cases}}
\label{formulation:3}
\setlength{\belowdisplayskip}{3pt}
\end{equation}
where $\lambda_{pre}$ is a hyper-parameter used to reduce the negative effect of excessive $\varnothing$ tokens,
$z_t$ symbolizes the $t$-th token of the target $z$,
and $\hat{p}_t^{i}$ is the $t$-th predicted probability distribution of the $i$-th slot using teacher forcing.
Meanwhile,
$\mathcal{L}^a(\theta, z)$ is defined in the similar way as $\mathcal{L}^p(\theta, z)$ with a hyper-parameter $\lambda_{abs}$.

\section{Preliminary Analyses}\label{subsection:cause-analyses}
Although \textsc{One2Set} has achieved competitive performance,
it still faces one major problem, i.e. \textit{$\varnothing$ token over-estimation}.
\textit{This occurs in such slots that produce $\varnothing$ tokens via the vanilla prediction while are able to generate correct keyphrases through the non-$\varnothing$ prediction\footnote{When performing the non-$\varnothing$ prediction, we remove $\varnothing$ token from the prediction vocabulary to generate a keyphrase.}.
}
For illustration,
we force all slots to generate non-$\varnothing$ predictions during inference,
where 14.6\% of slots can produce correct ones.
However, if we remove this restriction,
34.5\% of these slots directly output $\varnothing$ tokens,
revealing the over-estimation of $\varnothing$ token.
Such kind of miscalibration~\cite{DBLP:conf/icml/GuoPSW17, DBLP:journals/corr/abs-1903-00802} is a common drawback in neural network based models, which not only seriously hurts the generation quality of the \textsc{One2Set} paradigm, but also limits the users' trust towards it.

To understand the reasons behind this,
we use the commonly-used KP20k dataset \cite{meng-etal-2017-deep} to train a standard \textsc{SetTrans} model,
where the assigned targets to the slots of each instance are recorded during the last 80,000 training steps with an interval of 8,000 steps.
Here, we can obtain two crucial observations.

\begin{table}[t]
\footnotesize
\centering
\begin{tabular}{c|c|c}
\toprule
\multicolumn{1}{c|}{\multirow{1}{*}{\textbf{Target Type}}} & \textbf{Pre.KP Slots} & \textbf{Abs.KP Slots} \\
\midrule
\multirow{1}{*}{$\varnothing$}
& 72.4\% & 80.4\% \\
\multirow{1}{*}{Target KP}
& 27.6\% & 19.6\% \\
\bottomrule
\end{tabular}
\caption{The proportions of $\varnothing$ token and target keyphrases used as supervisory signals during training.
``KP'' means keyphrase.
``Pre.KP'' and ``Abs.KP'' represent present and absent keyphrases, respectively.\label{table:propotion-of-token}
}
\end{table}
\begin{table}[t]
\footnotesize
\centering
\setlength{\tabcolsep}{4.5mm}{
\begin{tabular}{c|c}
\toprule
\multicolumn{1}{c|}{\multirow{1}{*}{\textbf{Instance Type}}} & \textbf{All KP Slots} \\
\midrule
\multirow{1}{*}{Instance(\#OV-Slot=0)}  &   42.1\%\\
\multirow{1}{*}{Instance(\#OV-Slot=1)}  &   31.9\%  \\
\multirow{1}{*}{Instance(\#OV-Slot=2)}  &   15.5\%  \\
\multirow{1}{*}{Instance(\#OV-Slot$\geq$3)}  &  10.5\% \\
\bottomrule
\end{tabular}}
\caption{The proportions of instances involving different numbers of slots over-estimating $\varnothing$ token.
Instance(\#OV-Slot=$n$) means the instances containing $n$ slots over-estimating $\varnothing$ token.
Please note that the greater $n$,
the more severe $\varnothing$ token over-estimation.
}
\label{table:propotion-of-instance-overestimation}
\end{table}

\label{classimbalance_anylsis}
\paragraph{Observation 1:} \textit{Excessive $\varnothing$ tokens have been introduced as the padding tokens and served as supervisory signals in training data.}
\textsc{One2Set} models keyphrase generation in a parallel computation fashion,
therefore extensive padding $\varnothing$ tokens are used to make sure the fixed lengths of different samples.
Table \ref{table:propotion-of-token} shows the proportions of $\varnothing$ token and target keyphrases involved during the model training.
We can observe that on both present and absent keyphrase slots,
$\varnothing$ token accounts for the vast majority, exceeding 70\%.
In addition,
instances suffer from different degrees of $\varnothing$ token over-estimation. Table~\ref{table:propotion-of-instance-overestimation} shows the proportions of training instances grouped by the number of slots over-estimating $\varnothing$ token.
We can find that the instances (e.g. Instance(\#OV-Slot$\geq$1)) account for significant proportions,
and exist varying degrees of $\varnothing$ token over-estimation.

\paragraph{Observation 2:}
\textit{The $K$-step assignment mechanism is unstable and further increases the possibility of $\varnothing$ tokens being served as supervisory signals for some slots.}
In spite of the clever design of $K$-step assignment mechanism, it unstably provides different feasible target permutations to slots at the training time. We argue that this further widens the gap between the distribution of supervisory signals  and that of the ground-truth.

To illustrate this,
we classify the slots of each instance into three categories according to its target assignments:
1) Slot($\varnothing$),
each slot of this category is always assigned with $\varnothing$ tokens.
Apparently,
these slots hardly generate keyphrases after training;
2) Slot(Target KP), each slot of this category is always assigned with target keyphrases and thus it has high probability of generating a keyphrase;
3) Slot($\varnothing$+Target KP),
each slot is assigned with target keyphrases or $\varnothing$ tokens at different iterations during model training.
From Table \ref{table:propotion-of-slot},
we can observe that on both present and absent keyphrase slots,
the proportions of Slot($\varnothing$+Target KP) are quite high,
exceeding those of Slot(Target KP).
Quite evidently, the supervisory signals of slots in Slot($\varnothing$+Target KP) are unstable.
Those slots that should be labeled with Target KP are assigned with $\varnothing$ token,
further decreasing the probabilities of these slots generating keyphrases.

\begin{table}[t]
\label{table:propotion-of-kp-multislot}
\footnotesize
\centering
\setlength{\tabcolsep}{1.0mm}{
\begin{tabular}{c|c|c}
\toprule
\multicolumn{1}{c|}{\multirow{1}{*}{\textbf{Slot Type}}} & \textbf{Pre.KP Slots} & \textbf{Abs.KP Slots} \\
\midrule
\multirow{1}{*}{Slot($\varnothing$)}
& 61.2\% & 66.4\% \\
\multirow{1}{*}{Slot(Target KP)}
& 17.6\% & 9.3\% \\
\multirow{1}{*}{Slot($\varnothing$+Target KP)}
& 21.2\% & 24.4\% \\
\bottomrule
\end{tabular}}
\caption{The proportions of slots with different target assignments during the model training.
Slot($\varnothing$+Target KP) means the slots are assigned with $\varnothing$ tokens and target keyphrases alternatively at different iterations.
Note that the higher proportions of Slot($\varnothing$+Target KP),
the more slots contain unstable supervisory signals.}
\label{table:propotion-of-slot}
\end{table}

\begin{figure*}[t]
\centering
\footnotesize
\includegraphics[width=0.95\textwidth,
trim=400 980 770 160,clip]{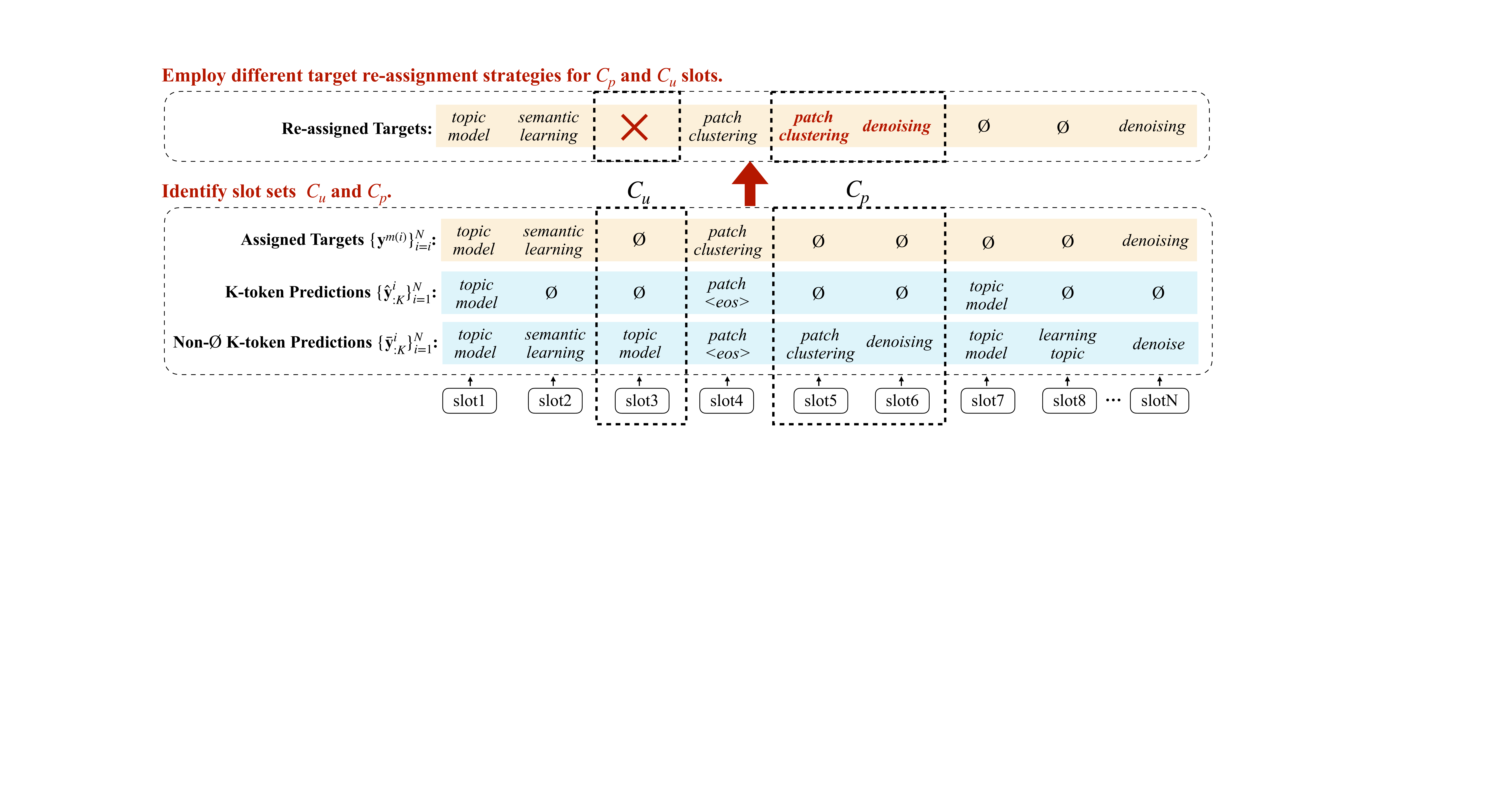}
\setlength{\abovecaptionskip}{5pt}
\caption{The procedure of target re-assignment involving two steps:
identifying the potential slot set $C_p$ and the unimportant one $C_u$,
and then employing different re-assignment operations to deal with them, respectively.
Here, ``\textcolor{red}{$\times$}'' represents assigning no supervisory signal to the slots of $C_u$, and following \citet{ye-etal-2021-one2set}, we set $K$ = 2.
}
\label{fig:re-assignment}
\end{figure*}

\section{\textsc{WR-One2Set}}
As discussed above, the parallelism and the training mechanism of \textsc{One2Set} bring the advantages of conditional independence,
but inherently lead to the miscalibration of the model.
Our principle is to maintain the primary advantages,
and meanwhile,
calibrating the model with lightweight strategies.
To this end,
we propose \textsc{WR-One2Set} that significantly extends the conventional \textsc{One2Set} paradigm in two training aspects,
including
an \textit{adaptive instance-level cost weighting strategy},
and
a \textit{target re-assignment mechanism}.

To facilitate subsequent descriptions,
we summarize all related formal definitions in Appendix, Table \ref{table:definations} for better understanding this paradigm.

\subsection{Adaptive Instance-Level Cost Weighting}
\textbf{Connection to Observation 1.}
As analyzed previously,
excessive $\varnothing$ tokens lead to the over-estimation of $\varnothing$ token.
Although \textsc{SetTrans} introduces hyper-parameters $\lambda_{pre}$ and $\lambda_{abs}$ to adjust the training loss of conventional \textsc{One2Set} paradigm,
such fixed hyper-parameters are still unable to deal with this issue well due to the different degrees of $\varnothing$ token over-estimation in different training instances.

We alternatively develop an adaptive instance-level cost weighting strategy to dynamically scale the losses corresponding to $\varnothing$ tokens,
alleviating the class imbalance of training data.
Concretely,
we first identify a set of slots,
denoted as $C_{!\varnothing}$,
where each slot is assigned with a keyphrase as supervisory signal.
Intuitively,
for each slot $i$ in $C_{!\varnothing}$,
the degree of $\varnothing$ token over-estimation is related to its two predicted probabilities using teacher forcing (See Section \ref{subsection:cause-analyses}):
1) $\hat{p}^{i}(y_0^{m(i)})$, symbolizing the predicted probability of the first token of assigned target,
and
2) $\hat{p}^{i}(\varnothing)$, denoting the predicted probability of $\varnothing$ token.
Thus,
we directly use the ratio between $\hat{p}^{i}(y_0^{m(i)})$ and $\hat{p}^{i}(\varnothing)$ to approximately quantify the degree of $\varnothing$ token over-estimation for training efficiency.
Furthermore,
we define this degree for each instance as
\begin{equation}
\setlength{\abovedisplayskip}{3pt}
\begin{aligned}
\lambda_{adp} =
\frac{1}{|C_{!\varnothing}|}\cdot\sum_{i \in C_{!\varnothing}} \min(\frac{
\hat{p}^{i}(y_0^{m(i)})
}{\hat{p}^{i}(\varnothing)
}, 1).
\label{formulation:4}
\end{aligned}
\setlength{\belowdisplayskip}{3pt}
\end{equation}
Note that for each slot $i$ in $C_{!\varnothing}$,
if its predicted probability $\hat{p}^{i}(y_0^{m(i)})$ is greater than $\hat{p}^{i}(\varnothing)$,
it is considered to have no $\varnothing$ token over-estimation,
and we directly limit its ratio to 1.

Finally,
we adjust the hyper-parameters $\lambda_{pre}$ and $\lambda_{abs}$ of Equation \ref{formulation:3} into $\lambda_{adp}$$\cdot$$\lambda_{pre}$
and $\lambda_{adp}$$\cdot$$\lambda_{abs}$ for each training instance, respectively.
Note that,
$\lambda_{adp}$
is dynamically updated during the training process,
and thus is more general for model training compared with fixed hyper-parameters.

\begin{table*}[ht]
\footnotesize
\centering
\setlength{\tabcolsep}{0.8mm}{
\begin{threeparttable}[width=0.8\textwidth]
\begin{tabular}{l|l|ll|ll|ll|ll|ll}
\toprule
\multicolumn{2}{c|}
{\multirow{2}{*}{\textbf{Model}}}&
				\multicolumn{2}{c|}{{\bf Inspec}}&\multicolumn{2}{c|}{{\bf NUS}}&\multicolumn{2}{c|}{{\bf Krapivin}}&\multicolumn{2}{c|}{{\bf SemEval}}&\multicolumn{2}{c}{{\bf KP20k}}\cr
				\multicolumn{2}{c|}{}&\textsc{$F_1@5$} &\textsc{$F_1@M$}&\textsc{$F_1@5$} &\textsc{$F_1@M$}&\textsc{$F_1@5$} &\textsc{$F_1@M$} & \textsc{$F_1@5$} &\textsc{$F_1@M$} & \textsc{$F_1@5$} &\textsc{$F_1@M$}  \cr
\midrule
\multicolumn{12}{c}{\textit{Existing Neural Keyphrase Generation Models}}\cr
\midrule
\multicolumn{2}{l|}{\textsc{CatSeq(R)}} &$0.225_{}$&$0.262_{}$&$0.323_{}$&$0.397_{}$&$0.269_{}$&$0.354_{}$&$0.242_{}$&$0.283_{}$&$0.291_{}$&$0.367_{}$  \cr
\multicolumn{2}{l|}{\textsc{CatSeq}} &$0.281_{5}$&$0.325_{6}$&$0.370_{7}$&$0.419_{10}$&$0.315_{8}$&$0.365_{5}$&$0.287_{14}$&$0.325_{15}$&$0.332_{1}$&$0.377_{1}$  \cr
\multicolumn{2}{l|}{\textsc{UniKeyphrase}} &0.260&0.288&0.415&0.443&\ \ \ ---&\ \ \ ---&0.302&0.322&0.347&0.352  \cr
\multicolumn{2}{l|}{\textsc{SetTrans}} &$0.285_3$&$0.324_3$&$0.406_{12}$&$0.450_7$&$0.326_{12}$&${0.364}_{12}$ & $0.331_{20}$ & $0.357_{13}$ & $0.358_5$ & $0.392_4$  \cr
\multicolumn{2}{l|}{\textsc{PromptKP}} &0.260&0.294&0.412&0.439&\ \ \ ---&\ \ \ ---&0.329&0.356&0.351&0.355  \cr
\midrule
\multicolumn{12}{c}{\textit{Our Models}}\cr
\midrule
\multicolumn{2}{l|}{\textsc{SetTrans}} &
			    $0.282_{2}$&$0.320_{2}$&$0.399_{5}$&$0.437_{8}$&$0.334_{7}$&$0.368_{4}$&$0.333_{6}$&$0.357_{4}$&$0.359_{2}$&$0.392_{2}$\cr
\multicolumn{2}{l|}{\textsc{SetTrans}(w/o $\lambda_{pre}$, $\lambda_{abs}$)} &
				$0.100_{4}$&	$0.148_{6}$&	$0.173_{18}$&	$0.258_{42}$&	$0.155_{9}$&	$0.280_{16}$&	0$.129_{7}$&	$0.188_{8}$&	$0.191_{7}$&	$0.321_{9}$\cr

\multicolumn{2}{l|}{\textsc{SetTrans(\#Slot=12)}} &$0.280_{1}$&$0.316_{2}$&$0.387_{9}$&$0.423_{4}$&$0.324_{5}$&$0.369_{3}$&$0.302_{6}$&$0.327_{4}$&$0.347_{3}$&$0.385_{3}$  \cr
\multicolumn{2}{l|}{\textsc{SetTrans(\#Slot=16)}} &$0.280_{7}$&$0.318_{6}$&$0.384_{7}$&$0.431_{10}$&$0.319_{3}$&$0.362_{1}$&$0.316_{15}$&$0.356_{17}$&$0.342_{2}$&$0.382_{2}$ \cr
\multicolumn{2}{l|}{\textsc{SetTrans(\#Slot=24)}} &$0.284_{13}$&$0.320_{18}$&$0.400_{1}$&$0.453_{1}$&$0.331_{1}$&$0.367_{5}$&$0.327_{18}$&$0.359_{11}$&$0.360_{1}$&${\bf 0.395}_{1}$  \cr
\multicolumn{2}{l|}{\textsc{SetTrans(\#Slot=28)}} &$0.277_{9}$&$0.317_{8}$&$0.402_{1}$&$\textbf{0.454}_{4}$&$0.338_{4}$&${\bf0.374}_{6}$&$0.315_{3}$&$0.349_{2}$&$0.355_{4}$&$0.392_{4}$  \cr

\multicolumn{2}{l|}{\textsc{SetTrans}(w/ \textsc{Batching})}
&$0.281_{7}$&$0.271_{6}$&$0.379_{4}$&$0.358_{4}$&$0.316_{2}$&$0.264_{7}$&$0.300_{8}$&$0.293_{6}$&$0.341_{2}$&$0.304_{3}$  \cr

\midrule

\multicolumn{2}{l|}{\textsc{Our Model}} &
			$\textbf{0.330}_{3}\ddag$&$\textbf{0.351}_{3}\ddag$&$\textbf{0.428}_{5}\ddag$&$0.452_{1}$&$\textbf{0.360}_{4}\ddag$&$0.362_{5}$&$\textbf{0.360}_{5}\ddag$&$\textbf{0.370}_{2}\ddag$&$\textbf{0.370}_{1}$&$0.378_{2}$ \cr

\bottomrule
\end{tabular}
\end{threeparttable}
\caption{Results of present keyphrase prediction.
Results shown in the upper part are directly cited from their corresponding papers.
The subscript denotes the corresponding standard deviation (e.g., 0.330$_3$ indicates 0.330±0.003).
$\ddag$ indicates significant at $p$<$0.01$ over \textsc{SetTrans} with 1,000 booststrap tests \cite{DBLP:books/sp/EfronT93}.\\
}\label{table:present-main-performance}}

\end{table*}
\begin{table*}[ht]
\footnotesize
\centering
\setlength{\tabcolsep}{0.5mm}{
\begin{threeparttable}[width=0.8\textwidth]
\begin{tabular}{l|l|ll|ll|ll|ll|ll}
\toprule
\multicolumn{2}{c|}
{\multirow{2}{*}{\textbf{Model}}}&
				\multicolumn{2}{c|}{{\bf Inspec}}&\multicolumn{2}{c|}{{\bf NUS}}&\multicolumn{2}{c|}{{\bf Krapivin}}&\multicolumn{2}{c|}{{\bf SemEval}}&\multicolumn{2}{c}{{\bf KP20k}}\cr
				\multicolumn{2}{c|}{}&\textsc{$F_1@5$} &\textsc{$F_1@M$}&\textsc{$F_1@5$} &\textsc{$F_1@M$}&\textsc{$F_1@5$} &\textsc{$F_1@M$} & \textsc{$F_1@5$} &\textsc{$F_1@M$} & \textsc{$F_1@5$} &\textsc{$F_1@M$}  \cr
\midrule	\multicolumn{12}{c}{\textit{Existing Neural Keyphrase Generation Models}}\cr
\midrule
\multicolumn{2}{l|}{\textsc{CatSeq(R)}} &0.004&0.008&0.016&0.028&0.018&0.036&0.016&0.028&0.015&0.032  \cr
\multicolumn{2}{l|}{\textsc{CatSeq}} &$0.010_{2}$&$0.019_{4}$&$0.028_{2}$&$0.048_{2}$&$0.032_{1}$&$0.060_{4}$&$0.020_{5}$&$0.023_{3}$&$0.023_{1}$&$0.046_{1}$  \cr
\multicolumn{2}{l|}{\textsc{UniKeyphrase}} &0.012&0.022&0.026&0.037&\ \ \ ---&\ \ \ ---&0.022&0.029&0.032&0.058  \cr

\multicolumn{2}{l|}{\textsc{SetTrans}} &$0.021_{1}$&$\textbf {0.034}_{3}$&$0.042_{2}$&$0.060_{4}$&$0.047_{7}$&$0.073_{11}$&$0.026_{3}$&$0.034_{5}$&$0.036_{2}$&$0.058_{3}$ \cr
\multicolumn{2}{l|}{\textsc{PromptKP}} &0.017&0.022&0.036&0.042&\ \ \ ---&\ \ \ ---&0.028&0.032&0.032&0.042  \cr

\midrule
\multicolumn{12}{c}{\textit{Our Models}}\cr
\midrule

\multicolumn{2}{l|}{\textsc{SetTrans}} &
				$0.020_{3}$&$0.031_{4}$&$0.044_{5}$&$0.061_{8}$&$0.050_{2}$&$0.073_{1}$&$0.030_{2}$&$0.037_{1}$&$0.038_{1}$&	$0.059_{1}$\cr
\multicolumn{2}{l|}{\textsc{SetTrans}(w/o $\lambda_{pre}$, $\lambda_{abs}$)} &$0.000_{0}$&	$0.000_{0}$&$0.002_{1}$&$0.003_{2}$&$0.004_{1}$&${0.008_{2}}$&$0.002_1$&$0.003_{2}$&$0.002_{1}$&$0.005_{1}$\cr

\multicolumn{2}{l|}{\textsc{SetTrans(\#Slot=12)}} &$0.016_{3}$&$0.027_{6}$&$0.042_{7}$&$0.065_{13}$&$0.047_{5}$&$0.073_{9}$&$0.024_{8}$&$0.031_{8}$&$0.033_{1}$&$0.057_{2}$  \cr

\multicolumn{2}{l|}{\textsc{SetTrans(\#Slot=16)}} &$0.018_{1\ }$&$0.030_{2\ }$&$0.040_{5\ }$&$0.060_{8\ }$&$0.045_{3\ }$&$\textbf{0.074}_{2\ }$&$0.023_{1\ }$&$0.031_{1\ }$&$0.034_{1\ }$&$0.057_{2\ }$ \cr

\multicolumn{2}{l|}{\textsc{SetTrans(\#Slot=24)}} &$0.019_{2}$&$0.029_{5}$&$0.044_{3}$&$0.061_{4}$&$0.046_{5}$&$0.073_{9}$&$0.026_{3}$&$0.035_{4}$&$0.038_{1}$&$0.059_{2}$  \cr

\multicolumn{2}{l|}{\textsc{SetTrans(\#Slot=28)}} &$0.016_{3}$&$0.026_{5}$&$0.044_{4}$&$0.063_{3}$&$0.043_{1}$&$0.070_{1}$&$0.021_{4}$&$0.027_{2}$&$0.032_{3}$&$0.054_{3}$  \cr
\multicolumn{2}{l|}{\textsc{SetTrans}(w/ \textsc{Batching})} &$0.023_{1}$&$0.030_{4}$&$0.050_{5}$&$0.067_{6}$&$0.049_{3}$&$0.059_{9}$&$0.034_{4}$&$0.038_{6}$&$0.045_{2}$&$0.058_{2}$  \cr
\midrule



\multicolumn{2}{l|}{\textsc{Our Model}} &
			$\textbf{0.025}_{2}$&$\textbf {0.034}_{4}$&$\textbf {0.057}_{5}\ddag$&$\textbf {0.071}_{3}\ddag$&$\textbf {0.057}_{1}\ddag$&$\textbf {0.074}_{2}$&$\textbf {0.040}_{3}\ddag$&$\textbf {0.043}_{5}\ddag$&$\textbf {0.050}_{1}\ddag$&$\textbf {0.064}_{2}$ \cr	
\bottomrule
\end{tabular}
\end{threeparttable}
\caption{Results of absent keyphrase prediction.}
\label{table:absent-main-performance}
}
\end{table*}

\subsection{Target Re-Assignment}\label{subsec:target_reassign}
\textbf{Connection to Observation 2.}
Due to the effect of $K$-step target assignment mechanism,
many slots are alternatively assigned with target keyphrases and $\varnothing$ tokens,
which decreases the probabilities of these slots generating correct keyphrases.

We propose a target re-assignment mechanism to alleviate this issue.
As shown in the upper part of Figure \ref{fig:re-assignment},
during the process of $K$-step target assignment,
we first record three kinds of phrases for each slot $i$:
1) $y^{m(i)}$,
the assigned target of the slot $i$;
2) $\hat{y}_{:K}^i$,
the first $K$ tokens of the vanilla prediction from the slot $i$.
Note that $\hat{y}_{:K}^i$ may be a $\varnothing$ token;
and
3) $\bar{y}_{:K}^i$,
the first $K$ tokens of the non-$\varnothing$ prediction from the slot $i$.

Here,
we mainly focus on the slots,
each of which is assigned with $\phi$ token as supervisory signals and its non-$\phi$ K-token prediction is consistent with some targets.
For such slot $i$,
if its non-$\varnothing$ $K$-token prediction $\bar{y}_{:K}^i$ is totally different from all $K$-token predictions $\{\hat{y}_{:K}^{i}\}_{i=1}^N$,
we consider it has the potential to generate a fresh keyphrase and boost the model performance.
Thus,
we include it into the \textbf{potential slot set} $C_p$.
By contrast,
if its $\bar{y}_{:K}^i$ has occurred in the set of $\{\hat{y}_{:K}^{i}\}_{i=1}^N$,
we regard it as an unimportant slot without effect on the model performance,
and add it into the \textbf{unimportant slot set} $C_u$.
Back to Figure \ref{fig:re-assignment},
we observe that the non-$\varnothing$ $K$-token prediction of ``slot3'' is ``\emph{topic model}'',
which is also the $K$-token prediction of ``slot1'' and ``slot7''.
Thus, ``slot3'' is an unimportant slot.
Meanwhile,
both ``slot5'' and ``slot6'' are potential slots.

Then,
as illustrated in the lower part of Figure \ref{fig:re-assignment},
we employ two target re-assignment operations to deal with the above two kinds of slots, respectively:
1) we re-assign each slot of $C_p$ with its best-matched target keyphrase,
so as to increase the probability of this slot generating the target keyphrase;
and
2)
we assign no target to each slot of $C_u$,
which alleviates the problem that the same target is assigned to different slots as supervisory signals.
In this way,
the training losses of slots in $C_u$ will be masked at this training iteration.
Let us revisit Figure \ref{fig:re-assignment},
we re-assign ``slot5'' with ``\textit{patch clustering}'', ``slot6'' with ``\textit{denoising}''
and no supervisory signal to ``slot3''.
Through the above process,
we can convert the original target assignment $m$ into a new one,
where we use the conventional training objective
(See Equation \ref{formulation:2}) adjusted with $\lambda_{adp}$ (See Equation \ref{formulation:3}) to train our model.

\section{Experiments}
\subsection{Setup}
\textbf{Datasets}.
We train various models and select the optimal parameters on the KP20k validation dataset \cite{meng-etal-2017-deep}.
Then,
we evaluate these models on five test datasets:
Inspec \cite{hulth-2003-improved},
NUS \cite{DBLP:conf/icadl/NguyenK07},
Krapivin \cite{krapivin2009large},
SemEval \cite{DBLP:conf/semeval/KimMKB10}, and KP20k.
As implemented in \cite{yuan-etal-2020-one,ye-etal-2021-one2set},
we perform data preprocessing including tokenization, lowercasing, replacing all digits with the symbol $\langle$\emph{digit}$\rangle$ and removing duplicated instances.

\paragraph{Baselines.}
We compare our \textsc{WR-One2Set} based model with the following baselines:
\begin{itemize}
\setlength{\itemsep}{0pt}
\setlength{\parsep}{0pt}
\setlength{\parskip}{0pt}
\item \textbf{\textsc{CatSeq(R)}} \cite{yuan-etal-2020-one}.
This is the most popular RNN-based model trained under the \textsc{One2Seq} paradigm, formulating keyphrase generation as a sequence-to-sequence generation task.

\item \textbf{\textsc{CatSeq}} \cite{ye-etal-2021-one2set}.
It is also trained under the \textsc{One2Seq} paradigm,
but utilizing Transformer as backbone.

\item
\textbf{\textsc{UniKeyphrase}} \cite{DBLP:conf/acl/WuLLNCZW21}.
This is a large-scale pre-trained language model trained to extract and generate keyphrases jointly.

\item \textbf{\textsc{SetTrans}} \cite{ye-etal-2021-one2set}.
It is our most important baselines. Besides,
we report the performance of
three \textsc{SetTrans} variants:
\textbf{\textsc{SetTrans}(w/o $\lambda_{pre}$, $\lambda_{abs}$)} that does not introduce any hyper-parameter to alleviate the negative effect of excessive $\varnothing$ tokens,
\textbf{\textsc{SetTrans(\#Slot=N})} that is equipped with $N/2$ and $N/2$ slots for present target keyphrases and absent ones, respectively,
and \textbf{\textsc{SetTrans}(w/ \textsc{Batching})} which sorts all training instances in the increasing order of target keyphrase numbers and uses batch-wise randomized order to keep the padding length optimized.

\item \textbf{\textsc{PromptKP}} \cite{wu2022fast}.
It firstly extracts keywords for automatic prompt construction,
and then uses a mask-predict-based approach to generate the final absent keyphrase constrained by prompt.
\end{itemize}

\paragraph{Implementation Details.}
We use Transformer-\emph{base} \cite{DBLP:conf/nips/VaswaniSPUJGKP17} to construct all models.
During training,
we choose the top 50,002 frequent tokens to form the predefined vocabulary.
We use the Adam optimizer with a learning rate of 0.0001,
and a batch size of 12.
During inference,
we employ greedy search to generate keyphrases.
To ensure a fair comparison with \textsc{SetTrans},
we also set both slot numbers for present and absent keyphrases as 10,
the target assignment step $K$ as 2,
$\lambda_{pre}$ as 0.2 and $\lambda_{abs}$ as 0.1, respectively.
Particularly,
we run all experiments three times with different random seeds and report the average results,
so as to alleviate the impact of the instability of model training.

\paragraph{Evaluation Metrics.}
Following previous studies
\cite{chen-etal-2020-exclusive,ye-etal-2021-one2set},
we use macro averaged $F_1@5$ and $F_1@M$ to evaluate the quality of both present and absent keyphrases.
When using $F_1@5$,
if the prediction number is less than five,
blank keyphrases are added to make the keyphrase number reach five.
Particularly,
we employ the Porter Stemmer\footnote{\url{https://github.com/nltk/nltk/blob/develop/nltk/stem/porter.py}}
to remove the identical stemmed keyphrases.

\subsection{Main Results}
Table \ref{table:present-main-performance} and Table \ref{table:absent-main-performance} show the prediction results on present and absent keyphrases, respectively.
We can draw the following conclusions:

First, our reproduced \textsc{SetTrans} achieves comparable performance to \citet{ye-etal-2021-one2set}.
Second,
when removing both $\lambda_{pre}$ and $\lambda_{abs}$ from \textsc{SetTrans},
its performance significantly drops,
showing that
the $\varnothing$ token over-estimation severely limits its full potential.
Third,
we observe no improvements with different number of slots.
Fourth, the commonly-used batching method for sequence generation is not beneficial for \textsc{SetTrans}.
Finally,
our model significantly surpasses all baselines.
These results strongly validate the effectiveness and generalization of our \textsc{WR-One2Set} paradigm.

\begin{table}[t]
\footnotesize
\centering
\renewcommand\arraystretch{1.0}
\setlength{\tabcolsep}{0.2mm}{
\begin{threeparttable}
\begin{tabular}{l|l|cc|cc}
\toprule
\multicolumn{2}{c|}
{\multirow{2}{*}{\textbf{Model}}}&
	\multicolumn{2}{c|}{{\bf In-domain}}&\multicolumn{2}{c}{{\bf Out-domain}}\cr
	\multicolumn{2}{c|}{}& $F_1@5$ & $F_1@M$ & $F_1@5$ & $F_1@M$
\cr
\midrule
\multicolumn{6}{c}{\textit{Present Keyphrase Prediction}}\cr
\midrule
\multicolumn{2}{l|}{\textsc{Our Model}} & \textbf{0.370} & 0.378 & \textbf{0.370} & {\bf0.384}
\cr
\multicolumn{2}{l|}{w/o \textsc{Re-assign} } & 0.368 & 0.375 &0.360 & 0.377
\cr
\multicolumn{2}{l|}{w/o \textsc{Weighting} } & 0.365 & \textbf{0.393} &0.340 & 0.374
\cr
\multicolumn{2}{l|}{\textsc{Re-Assign$\Rightarrow$Rand-Assign}}& 0.368 & 0.377 & 0.365 & 0.380
\cr
\midrule
\multicolumn{2}{l|}{ \textsc{SetTrans} } & 0.359 & 0.392 & 0.336 & 0.373
\cr
\midrule
\multicolumn{6}{c}{\textit{Absent Keyphrase Prediction}}\cr
\midrule
\multicolumn{2}{l|}{\textsc{Our Model}} & \textbf{0.050} & {\bf 0.064} & {\bf 0.043} & {\bf 0.055}
\cr
\multicolumn{2}{l|}{w/o \textsc{Re-assign} }& 0.047 & 0.062 & 0.042 & 0.053
\cr
\multicolumn{2}{l|}{w/o \textsc{Weighting} }& 0.043 & 0.063 & 0.039 & 0.052
\cr
\multicolumn{2}{l|}{\textsc{Re-Assign$\Rightarrow$Rand-Assign}}& 0.048 & 0.063 & 0.042 & 0.053
\cr
\midrule
\multicolumn{2}{l|}{ \textsc{SetTrans} } & 0.038 & 0.059 & 0.034 & 0.052
\cr
\bottomrule
\end{tabular}\caption{Ablation study on keyphrase predictions.}
\label{table:ablation-kp}
\end{threeparttable}}
\end{table}

\subsection{Ablation Study}
To better investigate the effectiveness of our proposed strategies on \textsc{WR-One2Set},
we report the performance of variants of our model on two test sets:
1) KP20k that is an \textbf{in-domain} one,
and
2) the combination of Inspec, NUS, Krapivin and SemEval,
which is \textbf{out-domain}.
Here, we mainly consider three variants:
1) w/o \textsc{Re-assign}, which removes the target re-assignment mechanism from our model;
and 2) w/o \textsc{Weighting}. It discards the adaptive instance-level cost weighting strategy;
and 3) \textsc{Re-Assign$\Rightarrow$Rand-Assign}. This variant randomly re-assigns targets to the slots in $C_p$.

As shown in Table \ref{table:ablation-kp},
when removing the target re-assignment mechanism,
we observe a performance degradation on keyphrase predictions.
Likewise,
the variant w/o \textsc{Weighting}
is obviously inferior to our model on most metrics.
Therefore,
we believe that our proposed strategies indeed benefit the generation of keyphrase set.

\subsection{Analyses of $\varnothing$ Token Over-Estimation}
We also compare various models according to the proportion of slots over-estimating $\varnothing$ tokens.
Here,
the proportion is the ratio between two slot numbers obtained from the whole training data:
one is the number of slots that directly output $\varnothing$ token via the vanilla prediction while generating correct keyphrases through the non-$\varnothing$ prediction;
and the other is the number of slots that generate correct keyphrases via the non-$\varnothing$ prediction.
Table \ref{table:proportion_overestimation_null} displays the results.
The proportions of \textsc{SetTrans} (w/o $\lambda_{pre}$, $\lambda_{abs}$) exceeds 70\%,
demonstrating the severe $\varnothing$ token over-estimation of the \textsc{One2Set} paradigm.
By comparison,
the proportions of \textsc{SetTrans} decrease,
validating the effectiveness of fixed hyper-parameters $\lambda_{pre}$ and $\lambda_{abs}$ on alleviating the class imbalance of training data.
Moreover,
whether adaptive instance-level cost weighting strategy or target re-assignment mechanism is used alone, the proportions of \textsc{SetTrans} can be further reduced.
Particularly, our model achieves the lowest proportions, proving that our strategies can complement each other.
\begin{table}[t]
\footnotesize
\centering
\setlength{\tabcolsep}{1.1mm}{
\begin{threeparttable}
\begin{tabular}{l|l|c|c}
\toprule
\multicolumn{2}{c|}
{\multirow{1}{*}{\textbf{Model}}} & \multicolumn{1}{c|}{{\bf In-domain}} & \multicolumn{1}{c}{{\bf Out-domain}}\cr
\midrule
\multicolumn{2}{l|}{ \textsc{SetTrans}(w/o $\lambda_{pre}$, $\lambda_{abs}$) } & 0.747&0.809\cr
\multicolumn{2}{l|}{ \textsc{SetTrans} }                       & 0.345 & 0.418 \cr
\multicolumn{2}{l|}{\textsc{SetTrans}(w/ \textsc{Re-assign})} & 0.301 & 0.386 \cr
\multicolumn{2}{l|}{\textsc{SetTrans}(w/ \textsc{Weighting})} & 0.240 & 0.308 \cr
\multicolumn{2}{l|}{\textsc{Our Model}}                        & \textbf{0.211} & \textbf{0.263}   \cr
\bottomrule
\end{tabular}
\caption{
The proportions of slots over-estimating $\varnothing$ token.
\label{table:proportion_overestimation_null}}
	\setlength{\belowcaptionskip}{0pt}
\end{threeparttable}}
\end{table}
\begin{figure}[t]
   \centering
	\footnotesize
	\subfigure[\textsc{SetTrans}]{
		\includegraphics[width=0.210\textwidth,
	trim=5 0 5 0,clip]{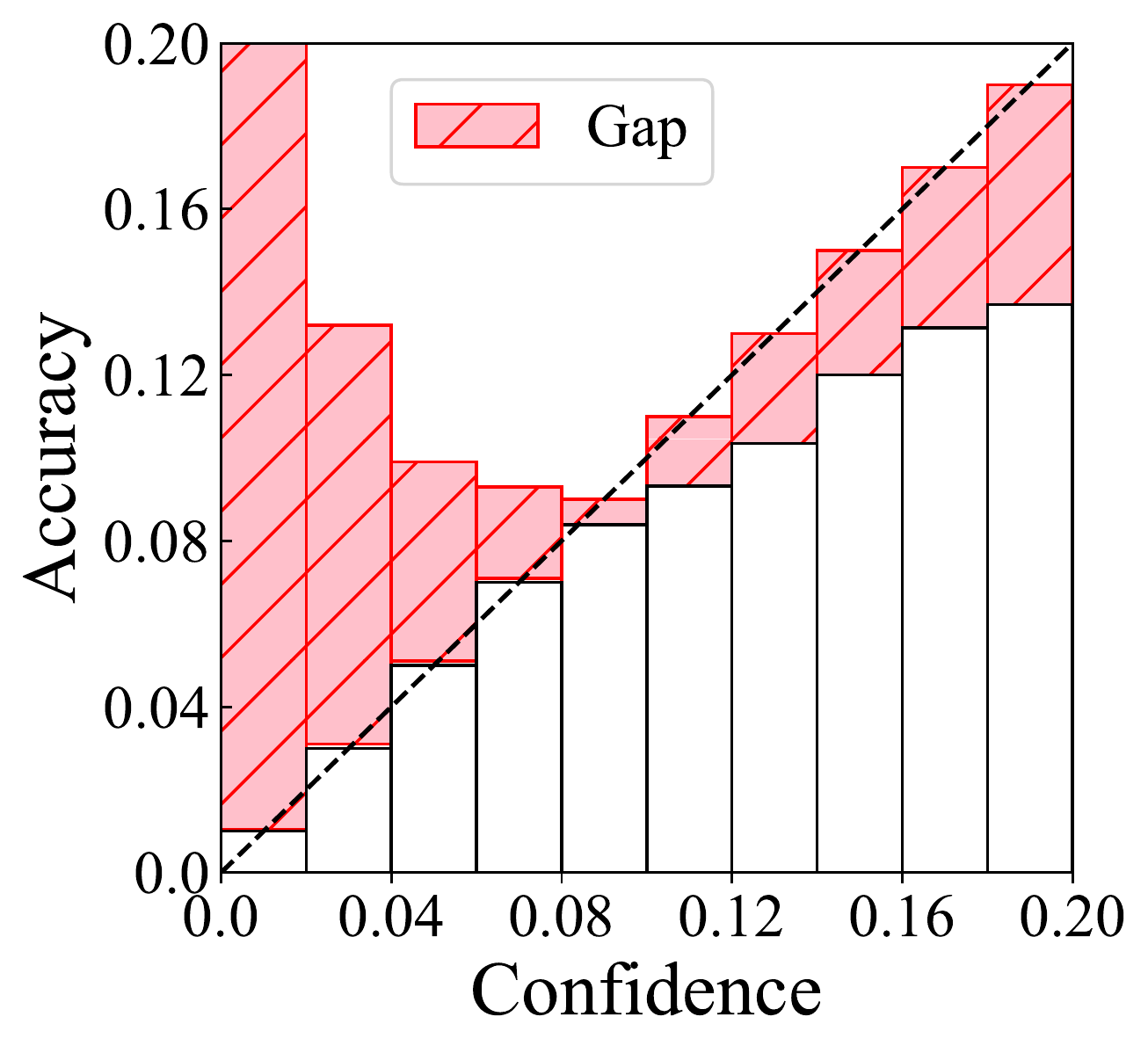}
	}
   \subfigure[\textsc{Our Model}]{
		\includegraphics[width=0.195\textwidth,
	trim=5 0 5 0,clip]{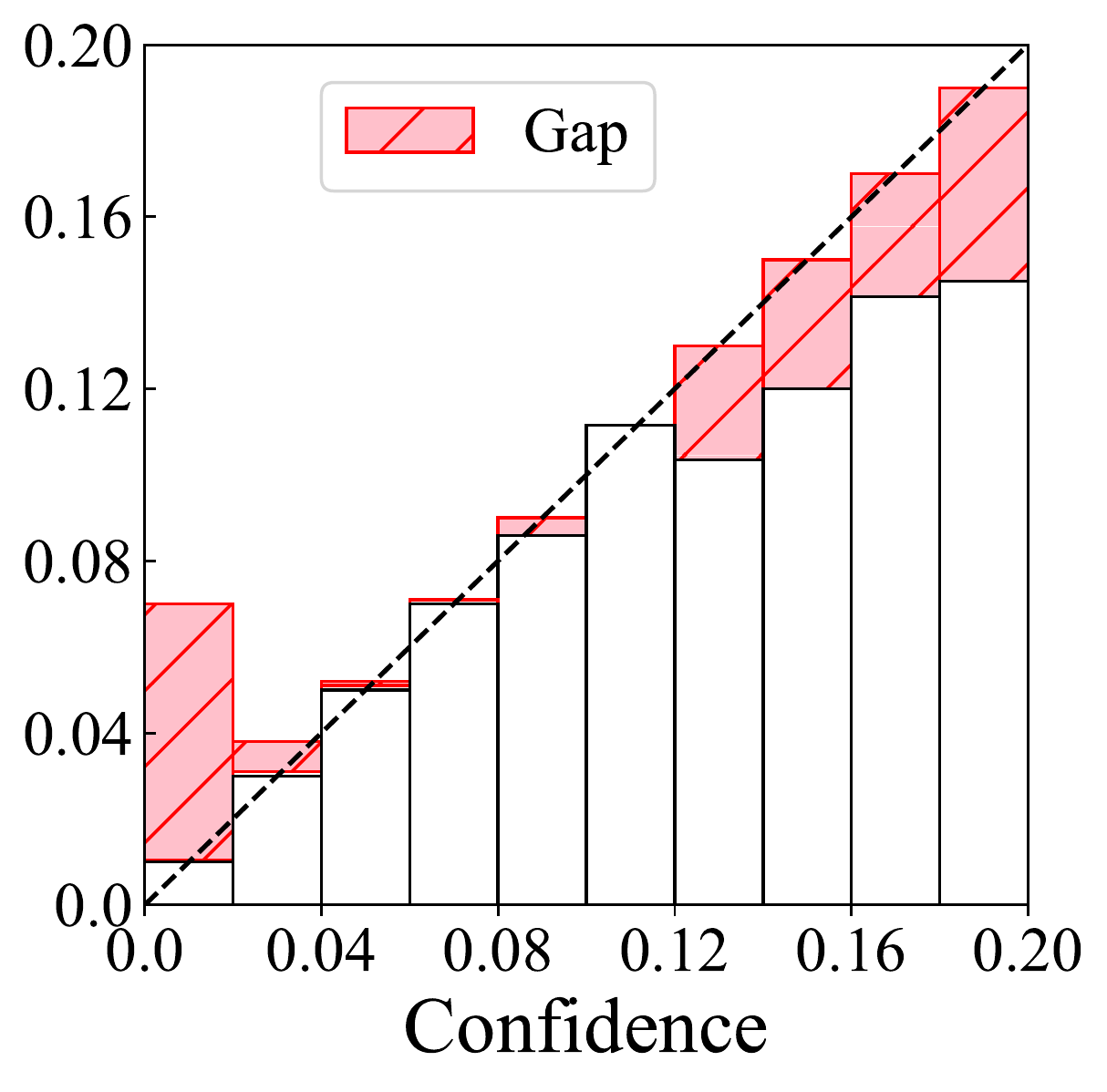} }
	\setlength{\abovecaptionskip}{0pt}
\caption{Reliability diagrams of \textsc{SetTrans} and our model on the in-domain test set.
``Gap'' (areas marked with slash) denotes the difference between the prediction confidence and the prediction accuracy.
Smaller gaps denote better calibrated outputs.
\label{fig:reliability_diagram}}
\end{figure}

Besides,
following \citet{DBLP:conf/icml/GuoPSW17},
we show the reliability diagram of \textsc{SetTrans} and our model in Figure \ref{fig:reliability_diagram}.
It displays the relationship between the prediction confidence (the predicted probability of model) and the prediction accuracy within the confidence interval [0, 0.2].
Especially,
the predictions within the confidence interval [0, 0.2]
account for 69.8\% of all predictions.
Please note that,
if a model is well-calibrated,
the gap between the confidence and the accuracy will be small.
Overall, the gap of our model is less than that of \textsc{SetTrans},
which demonstrates that our proposed strategies can calibrate the predictions with low confidence.

\subsection{Diversity of Predicted Keyphrases}
\begin{table}[t]
\footnotesize
\centering
\renewcommand\arraystretch{1.2}
\setlength{\tabcolsep}{0.3mm}{
\begin{threeparttable}[width=0.5\textwidth]
\begin{tabular}{l|l|ccc|ccc}
\toprule
			    \multicolumn{2}{c|}
				{\multirow{2}{*}{\textbf{Model}}}&
				\multicolumn{3}{c|}{{\bf In-domain}}&\multicolumn{3}{c}{{\bf Out-domain}}\cr
				\multicolumn{2}{c|}{}&\#Pre &\#Abs&Dup& \#Pre&\#Abs&Dup\cr
				\cline{1-8}
				\multicolumn{2}{l|}{\textsc{Oracle}} &3.31&1.95&-&5.53&3.51&-  \cr
				\cline{1-8}
				\multicolumn{2}{l|}{\textsc{CatSeq(R)}} &{\bf3.71}&0.55&0.39&3.46&0.72&0.54   \cr
				\multicolumn{2}{l|}{\textsc{CatSeq} }&4.64&1.16&0.26&4.34&1.28&0.38  \cr
				\multicolumn{2}{l|}{\textsc{SetTrans}} & 5.10 & {\bf 2.01}& {\bf0.08}&4.62 & 2.18 & {\bf0.08}   \cr
				\cline{1-8}	
            \multicolumn{2}{l|}{\textsc{SetTrans}(w/ \textsc{Re-assign})} & 5.40 & 2.64  & 0.10 & 4.83 &  2.72 & 0.09 \cr
            \multicolumn{2}{l|}{\textsc{SetTrans}(w/ \textsc{Weighting})} & 6.19 & 3.12  & 0.11 & {\bf5.70} &  {\bf3.56} & 0.09 \cr
            \multicolumn{2}{l|}{\textsc{Ours Model}} & 6.35 & 3.26 & 0.10 & 5.94 &  3.60 &0.10 \cr

			\bottomrule

\end{tabular}
\end{threeparttable}
\caption{Numbers and duplication ratios of predicted keyphrases on test datasets.
``\textsc{Oracle}'' refers to the average number of target keyphrases.}
\label{table:diversity-of-kp}
}
\end{table}

Follow previous studies \cite{chen-etal-2020-exclusive,ye-etal-2021-one2set},
we report the average numbers of unique present and absent keyphrases,
and the average duplication ratios of all predicted keyphrases,
so as to investigate the ability of our model in generating diverse keyphrases,
Table \ref{table:diversity-of-kp} reports the results.
As expected, our model generates more keyphrases than previous models and achieves a slightly higher duplication ratio than \textsc{SetTrans}, however, significantly lower than \textsc{One2Seq}-based models.
Note that compared to \textsc{SetTrans}, the $F_1@5$ and $F_1@M$ scores of our model are significantly improved, which demonstrates that our model performs much better on keyphrase generation.

\subsection{Analyses of Target Re-Assignment}
\begin{table}[t]
\small
\centering
\setlength{\tabcolsep}{2.5mm}{
\begin{threeparttable}
\begin{tabular}{c|c|c}
\toprule
\footnotesize
\bf{Slot Type} & \bf{\textsc{SetTrans}} & \bf{\textsc{Our Model}} \cr
\midrule
\multicolumn{3}{c}{\textit{Present Keyphrase Prediction}}\cr
\midrule
\multirow{1}{*}{Slot($\varnothing$)}           & 61.2\% & 63.5\% (+2.3\%)\\
\multirow{1}{*}{Slot(Target KP)}               & 17.6\% & 21.8\% (+4.2\%)\\
\multirow{1}{*}{Slot($\varnothing$+Target KP)} & 21.2\% & 14.7\% (-6.5\%)\\
\midrule
\multicolumn{3}{c}{\textit{Absent Keyphrase Prediction}}\cr
\midrule
\multirow{1}{*}{Slot($\varnothing$)}           & 66.4\% & 69.6\% (+3.2\%) \\
\multirow{1}{*}{Slot(Target KP)}               & 9.3\%  & 12.6\% (+3.3\%) \\
\multirow{1}{*}{Slot($\varnothing$+Target KP)} & 24.4\% & 17.8\% (-6.6\%) \\
\bottomrule
\end{tabular}
\caption{The proportions of slots with different target assignments for keyphrase predictions.
}
\label{table:proportions_stabled}
\end{threeparttable}
}
\end{table}

Here,
we still focus on the assigned targets during the model training mentioned at the beginning of Section \ref{subsection:cause-analyses} and conduct two types of analyses to better understand the effects of our mechanism.

First,
we count the proportions of $\varnothing$ tokens in assigned targets.
Specially,
the assigned $\varnothing$ tokens accounts for 72.4\% and 80.4\% on present and absent keyphrase slots, respectively,
but decrease to 67.6\% and 72.3\% in our model.
Second,
as implemented in Section \ref{subsection:cause-analyses},
we still classify the instance slots into three categories and report their proportions in Table \ref{table:proportions_stabled}.
We can find the proportions of Slots($\varnothing$+KP),
where slots are assigned with target keyphrase and $\varnothing$ token at different iterations of model training, sharply decline.
Besides,
for each slot,
we use the entropy of target assignment distribution to measure the stability of its supervisory signals.
Furthermore,
we average the entropy values of all slots to quantify the stability of supervisory signals for each instance.
Consequently,
we find that the entropy decreases in 68.2\% of instances,
increases in 26.3\% of instances,
and remain unchanged in 5.5\% of instances.
These results indicate that our target re-assignment mechanism indeed not only reduces excessive target $\varnothing$ tokens,
but also alleviates the instability of supervisory signals.

\section{Conclusion}
In this paper,
we in-depth analyze the serious calibration errors of the \textsc{One2set} paradigm and point out its underlying reasons.
To deal with this issue,
we then significantly extend the conventional \textsc{One2Set} into the \textsc{WR-One2Set} paradigm with an adaptive instance-level cost weighting strategy and a target re-assignment mechanism.
Extensive experiments verify the effectiveness and generality of our extended paradigm.

In the future,
we plan to further refine our \textsc{WR-One2Set} paradigm by considering the semantic relation between keyphrases.
Besides,
we will improve our model by introducing variational neural networks, which have been successfully applied in many NLP tasks \cite{
DBLP:conf/emnlp/ZhangXSDZ16,
DBLP:conf/emnlp/ZhangXSLJDZ16,
DBLP:conf/aaai/SuWXLHZ18,
DBLP:journals/isci/SuWZWQX18,
DBLP:conf/acl/LiangMZXCSZ22}.
Finally,
we will leverage the abundant knowledge from pre-trained models to further enhance our model.

\section*{Limitations}
As mentioned above,
serious $\varnothing$ token over-estimation problem exists in \textsc{One2Set} paradigm, leading to a miscalibrated model.
To solve this problem,
we propose several strategies based on conventional \textsc{One2Set} using the same fixed hyper-parameters as \citet{ye-etal-2021-one2set}.
However,
hyper-parameter selection
is a labor-intensive, manual, time-consuming process and affects generation performance deeply.
Thus,
our future work will focus on exploring a parameter-free method.
Besides,
despite achieving impressive performance,
our \textsc{WR-One2Set} paradigm is only conducted based on the Transformer,
so that it is essential to leverage the abundant knowledge from pre-trained models for better document modeling and keyphrase generation.

\section*{Acknowledgements}
The project was supported by National Natural Science Foundation of China (No. 62276219, No. 62036004),  Natural Science Foundation of Fujian Province of China (No. 2020J06001), and Youth Innovation Fund of Xiamen (No. 3502Z20206059). We also thank the reviewers for their insightful comments.

\bibliography{anthology,custom}
\bibliographystyle{acl_natbib}
\newpage
\appendix
\section{Appendix}
\subsection{Example}\label{subsec:example}
\begin{minipage}[t]{1.0\textwidth}
\begin{center}
\footnotesize
\begin{threeparttable}[width=1\textwidth]
\begin{tabular}{p{15cm}}
\toprule
\textbf{\emph{Input Document:}} an image \underline{topic model} for image \underline{denoising}. \underline{topic model} is a powerful tool for the basic document or image processing tasks. in this study,
we introduce a novel image \underline{topic model}, called latent patch model (lpm), which is a generative bayesian model and assumes that the image and pixels are connected by a latent patch layer.
based on the lpm, we further propose an image \underline{denoising} algorithm namely multiple estimate lpm (melpm). unlike other works, the proposed \underline{denoising} framework is totally implemented on the latent patch layer, and it is effective for both gaussian white noises and impulse noises. experimental results demonstrate that lpm performs well in representing images... \cr
\midrule
\textbf{\emph{Keyphrases:\  }}\underline{topic model}; \underline{denoising}; patch clustering; semantic learning \\
\bottomrule
\end{tabular}
\caption{
An example of keyphrase generation.
The underlined phrases are \textit{present keyphrases} that appear in the document,
and
other phrases are \textit{absent keyphrases} that do not match any contiguous subsequence of the document.
}
\end{threeparttable}
\end{center}

\label{table:present-cn-main-performance}

\end{minipage}

\subsection{Formal Definitions}\label{subsec:defination}
\begin{minipage}[t]{1.0\textwidth}

\footnotesize
\centering
\setlength{\tabcolsep}{5mm}{
\begin{threeparttable}[width=1\textwidth]
\begin{tabular}{cp{120mm}}
\toprule
\multicolumn{1}{c|}
{\multirow{1}{*}{\textbf{Symbol}}} & \multicolumn{1}{l}{\textbf{Definition}}\cr

\midrule
\multicolumn{1}{c|}{$\varnothing$}&{A special token representing no corresponding keyphrase.}\cr
\hline

\multicolumn{1}{c|}{$N$}&{The predefined number of slots generating keyphrases or $\varnothing$ tokens in parallel.}\cr
\hline

\multicolumn{1}{c|}{\multirow{2}{*}{$K$}}&{The predefined number of tokens generated from each slot for the conventional target assignment mechanism.}\cr
\hline


\multicolumn{1}{c|}{$\mathbf{P}^i$}&{The predicted probability distributions of the slot $i$.}\cr
\hline

\multicolumn{1}{c|}{\multirow{2}{*}{$\hat{p}^i(*)$}}&{The predicted probability of a keyphrase from the slot $i$ using teacher forcing. Specially, the $j$-th token predictive probability of $\hat{p}^i(*)$ is denoted as $\hat{p}^i_j(*)$.}\cr
\hline

\multicolumn{1}{c|}{$M(N)$} & {The set of all $N$-length target index permutations.}\cr
\hline

\multicolumn{1}{c|}{\multirow{3}{*}{$m$}}&{
The optimal permutation of $M(N)$.
It can be considered as a mapping function from the slot $i$ to the target index $m(i)$.
Particularly,
we use $m^p$ and $m^a$ to denote the optimal permutations for present and absent keyphrases, respectively.
}\cr
\hline

\multicolumn{1}{c|}{$y^{m(i)}$}&{The assigned target of the slot $i$.}\cr
\hline

\multicolumn{1}{c|}{\multirow{2}{*}{$\lambda_{pre}$, $\lambda_{abs}$}} & {Two predefined hyper-parameters used to reduce the negative effect of excessive $\varnothing$ tokens for present and absent keyphrase predictions, respectively. }\cr
\hline

\multicolumn{1}{c|}{\multirow{2}{*}{$\lambda_{adp}$}}&{The degree of $\varnothing$ token over-estimation for each instance,
which is leveraged to dynamically scale the losses corresponding to $\varnothing$ tokens in our paradigm.}\cr
\hline

\multicolumn{1}{c|}{\multirow{3}{*}{$\mathcal{L}$($\theta$)}}&{The training loss of the whole model with parameters $\theta$.
Moreover,
we use $\mathcal{L}^p(\theta, z)$ to denote the present keyphrase training loss on the assigned target $z$.
The absent keyphrase training loss $\mathcal{L}^a(\theta, z)$ is defined in a similar way.
}\cr
\hline

\multicolumn{1}{c|}{$\hat{y}_{:K}^i$} & {The first $K$ tokens of the prediction from the slot $i$ via the vanilla prediction.}\cr
\hline

\multicolumn{1}{c|}{\multirow{2}{*}{$\bar{y}_{:K}^i$}}&{The first $K$ tokens of the prediction from the slot $i$ through the non-$\varnothing$ prediction,
where $\varnothing$ token is removed from the prediction vocabulary.}\cr
\hline

\multicolumn{1}{c|}{$C_{!\varnothing}$} & {The set of slots,
each of which is assigned with a keyphrase as supervisory signal.}\cr
\hline

\multicolumn{1}{c|}{\multirow{2}{*}{$C_{p}$}} & {The set of potential slots,
where each slot has the potential to generate a fresh keyphrase, boosting the performance of the model.
}\cr
\hline

\multicolumn{1}{c|}{$C_{u}$} & {The unimportant slot set,
where each slot has no effect on the model performance.
}\cr
\bottomrule
\end{tabular}
\caption{Formal Definitions}
\end{threeparttable}

\label{table:definations}
}
\end{minipage}
\end{document}